\pdfoutput=1
\documentclass[11pt]{article}

\usepackage[final]{acl}
\usepackage{times}
\usepackage{latexsym}

\usepackage[T1]{fontenc}

\usepackage[utf8]{inputenc}

\usepackage{microtype}

\usepackage{inconsolata}

\usepackage{graphicx}
\usepackage{amsmath}
\usepackage{amsfonts} 
\usepackage{amssymb}  
\usepackage{booktabs}
\usepackage{multirow}
\usepackage{graphicx}
\usepackage{xspace}
\usepackage{colortbl}
\usepackage{tcolorbox}
\usepackage{xcolor}
\usepackage{graphicx}
\usepackage{subfigure} 
\usepackage[table]{xcolor}
\definecolor{Highlight}{rgb}{0.92,0.94,1}
\definecolor{darkgreen}{rgb}{0,0.6,0} 

\title{GIFT: Guided Fine-Tuning and Transfer for Enhancing Instruction-Tuned Language Models}

\newcommand{\mname}{{\text GIFT}\xspace}

\author{%
    Zhiwen Ruan$^{1}$, \
    Yichao Du$^{2}$, \
    Jianjie Zheng$^{1}$,\
    Longyue Wang$^{2}$\\
    \textbf{Yun Chen}$^{3}$,
    \textbf{Peng Li}$^{4}$, 
    \textbf{Jinsong Su}$^{5}$,
    \textbf{Yang Liu}$^{4}$,
    \textbf{Guanhua Chen}$^{1}\thanks{\ \ Corresponding author.}$ \\
    $^1$Southern University of Science and Technology, 
    $^2$Alibaba Group \\
    $^3$Shanghai University of Finance and Economics,
    $^4$Tsinghua University,
    $^5$Xiamen University  \\
}

\begin{document}
\maketitle
\begin{abstract}
A promising paradigm for adapting instruction-tuned language models is to learn task-specific updates on a pretrained base model and subsequently merge them into the instruction-tuned model. However, existing approaches typically treat the instruction-tuned model as a passive target that is only involved at the final merging stage, without guiding the training process.
We propose \textbf{GIFT} (\textbf{G}u\textbf{i}ded \textbf{F}ine-Tuning and \textbf{T}ransfer), a simple and efficient framework that incorporates guidance from the instruction model into task adaptation. GIFT fine-tunes a low-rank adapter on the pretrained base model using confidence signals derived from the instruction-tuned model. The learned adapter is then merged into the instruction-tuned model, yielding task-specialized models that preserve general instruction-following behavior.
We evaluate GIFT on mathematical and knowledge-intensive benchmarks across multiple model families and scales. Results show that GIFT consistently outperforms direct fine-tuning and representative transfer-based baselines, while maintaining robust generalization and favorable test-time scaling behavior.
\end{abstract}

\section{Introduction}

Instruction-tuned open-source large language models (LLMs), such as Qwen and Llama, have demonstrated strong instruction-following and zero-shot generalization capabilities, driven by the scaling of model parameters and training data~\citep{llama3modelcard,qwen2.5}. These models are produced through expensive and carefully designed post-training pipelines, resulting in parameters that are finely balanced across general reasoning ability~\citep{ruan2025unveilingovermemorizationfinetuningllms,yang2026incoder,wang2026spposequencelevelppolonghorizon}, instruction-following behavior, and robustness~\citep{brown2020language,fedus2022switch,achiam2023gpt,wang-etal-2025-milora}. Consequently, adapting instruction-tuned models to specific downstream tasks remains challenging~\citep{ruan-etal-2025-layalign}, as naive fine-tuning can easily disrupt this balance and lead to instability or performance degradation~\citep{wu2025shadowfttuninginstructmodel}.

A common alternative is to perform supervised fine-tuning on the pretrained base model and then transfer the learned task-specific updates back to the instruction-tuned model~\citep{fleshman2024readaptreverseengineeredadaptation,lin-etal-2025-efficient,cao2025paramdelta}. Representative methods such as Shadow-FT~\citep{wu2025shadowfttuninginstructmodel} and Chat Vector~\citep{huang-etal-2024-chat} follow this paradigm by training adapters on the base model and merging them into the instruction model, while Re-Adapt and its variants further explore linear combinations of base parameters, instruction offsets, and task-adapted updates~\citep{fleshman2024readaptreverseengineeredadaptation}. Despite their effectiveness in certain settings, these approaches treat the instruction-tuned model as a passive target that only participates at the final merging stage, without influencing how task-specific knowledge is learned.

In this work, we argue that the instruction-tuned model can play a more active role in task adaptation by providing guidance during the learning process itself, rather than only participating in post-hoc merging. Based on this insight, we propose \textbf{GIFT} (\textbf{G}u\textbf{i}ded \textbf{F}ine-Tuning and \textbf{T}ransfer), a simple and efficient framework that incorporates signals from the instruction model into task adaptation.
Specifically, GIFT fine-tunes a low-rank adapter on the pretrained base model using pre-computed confidence signals from the instruction model. This guidance mechanism effectively redistributes the learning signal by suppressing updates from uncertain predictions and focusing optimization on regions consistent with the instruction model's alignment. Finally, the learned adapter is merged into the instruction-tuned model, yielding a task-specialized model that inherits new capabilities while maintaining robust merge compatibility.

We evaluate GIFT on a range of mathematical and knowledge-intensive tasks, including mathematics and medical QA, across multiple model families and scales. Empirical results demonstrate that GIFT consistently outperforms direct fine-tuning and existing merge-based baselines, while maintaining generalization and instruction-following capabilities. On mathematical benchmarks (Math500, Minerva Math, OlympiadBench, AIME 2024, and AMC 2023), GIFT improves accuracy by an average of 5.2\% over Llama3.1-8B-Instruct~\citep{llama3modelcard}, and on medical QA tasks~\citep{pal2022medmcqa,jin2021disease}, it achieves a 6.2\% gain. Overall, GIFT provides a principled and practical framework for adapting instruction-tuned language models, demonstrating that task-specific capabilities can be effectively acquired through guided fine-tuning on the base model and transferred back to the instruction model. \footnote{Our code is publicly available at \url{https://github.com/sustech-nlp/gift}.}

\section{Related Work}

\subsection{Model Arithmetic}
Foundational studies establish that the parameter space of neural networks supports arithmetic manipulation~\citep{izmailov2019averagingweightsleadswider, wortsman2022modelsoups}, a property formalized by Task Vectors~\citep{ilharco2023editing} to merge or unlearn capabilities via vector addition. To mitigate performance degradation arising from parameter interference, recent techniques employ sparsification strategies. TIES-Merging~\citep{yadav2023tiesmerging} eliminates redundancy by pruning low-magnitude updates and resolving sign conflicts, while DARE~\citep{yu2024languagemodelssupermario} stochastically drops redundant delta parameters followed by rescaling. 
It is important to note that these merging techniques address a fundamentally different problem from GIFT. TIES and DARE operate in a purely post-hoc manner, assuming task-specific updates are already learned and focusing on how to combine them with minimal interference. 
In contrast, GIFT targets the learning stage itself by shaping optimization on the base model using guidance from the instruction model. As a result, these methods are complementary rather than competing, and can be directly integrated into the GIFT pipeline if desired.

\subsection{Enhanced Instruct Model}
Recent research exploits the linear arithmetic property of model weights to enhance capabilities. RE-Adapt~\citep{fleshman2024readaptreverseengineeredadaptation} and Chat Vector~\citep{huang-etal-2024-chat} pioneered this by extracting alignment vectors—defined as the difference between instruct and base weights—and grafting them onto domain-adapted or multilingual backbones to transfer instruction-following skills. Extending this to model evolution, Param$\Delta$\citep{cao2025paramdelta} and Fine-tuning Transfer\citep{lin-etal-2025-efficient} demonstrate that these ``diff vectors'' can be propagated across model versions (e.g., Llama 3 to 3.1) to instantly replicate capabilities. Most recently, Shadow-FT~\citep{wu2025shadowfttuninginstructmodel} addresses the instability of direct instruction tuning by learning updates on the base model and transferring them to the instruct version, preserving alignment while incorporating task knowledge.

While effective, these approaches treat the instruction model merely as a passive recipient for weight merging. In contrast, GIFT uses the instruction model to provide offline token-level confidence signals during base-model adaptation. By using these signals to weigh the base model's loss, GIFT produces adapters that are more compatible with the instruction model upon merging.

\section{Method}
\label{sec:method}

\begin{figure*}[t]
  \includegraphics[width=\linewidth]{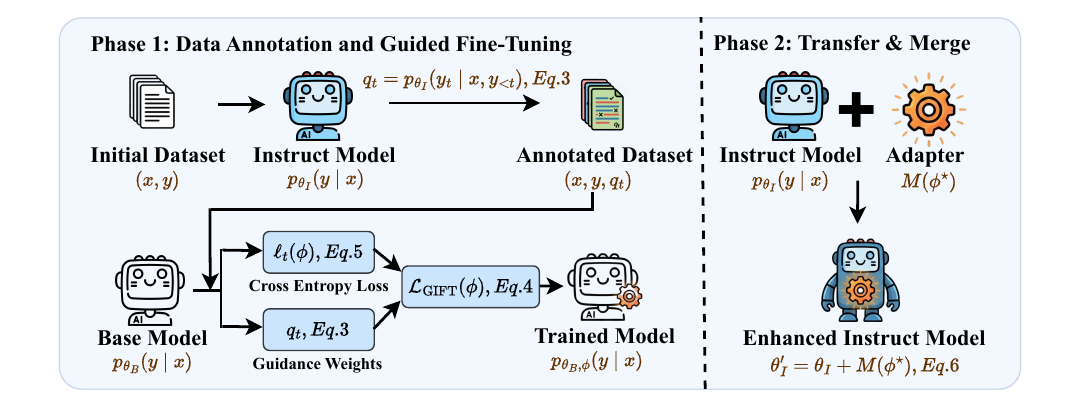}
  \caption{Overview of \mname. \textbf{Phase~1: Data Annotation and Guided Fine-Tuning.} 
The instruction-tuned model is first used to annotate the initial training dataset with confidence scores, producing an annotated dataset.
The base model is then fine-tuned with a low-rank adapter using these confidence-guided supervision signals.
\textbf{Phase~2: Transfer and Merge.} 
After training, the learned adapter is merged into the instruction-tuned model, yielding a task-enhanced instruction model without directly fine-tuning it.}
  \label{fig:main_fig}
\end{figure*}

\subsection{Setup and Objective}

Let $\mathcal{D} = \{(x,y)\}$ denote a supervised dataset. We denote the pretrained base model as $p_{\theta_B}(y \mid x)$ and the instruction-tuned model as $p_{\theta_I}(y \mid x)$. Task adaptation is performed by training a low-rank adapter $\phi$ on the base model.

Prior work, such as Shadow-FT~\citep{wu2025shadowfttuninginstructmodel} and Chat Vector~\citep{huang-etal-2024-chat}, suggests that updates can be transferred from the base model to the instruction model, as they share the same backbone and lie in a nearby region of parameter space:
\begin{equation}
    \theta_I = \theta_B + \Delta_I.
\end{equation}
Leveraging this property, task-specific updates learned on the base model (denoted $M(\phi)$) can be directly applied to the instruction model:
\begin{equation}
    \theta_I' = \theta_I + M(\phi),
    \label{eq:merge}
\end{equation}
where $M(\cdot)$ represents the merge operator (e.g., LoRA merging).

However, existing approaches typically treat the instruction model $\theta_I$ as a passive target, only consulted at the final merging stage. As a result, the adapter $\phi$ is trained using standard objectives that treat all tokens equally, overlooking the useful confidence information embedded in $\theta_I$.  
We instead use supervision signals from $\theta_I$ to measure token importance during base-model adaptation, allowing us to distinguish between valuable task knowledge and potential noise.  
Our objective is to leverage this guidance to learn a higher-quality adapter $\phi$, which, when merged via Eq.~\ref{eq:merge}, produces a more effective adapted model $\theta_I'$.

\subsection{Guided Fine-Tuning}

To train the adapter $\phi$ within the merge-based framework described above, prior methods typically rely on standard supervised fine-tuning (SFT)~\citep{wu2025shadowfttuninginstructmodel, huang-etal-2024-chat}.
This conventional approach optimizes the negative log-likelihood on the base model:
\begin{align*}
\mathcal{L}_{\text{SFT}}
&= \mathbb{E}_{(x,y) \sim \mathcal{D}} \Big[ 
\sum_{t=1}^{T} -\log p_{\theta_B,\phi}(y_t \mid x, y_{<t}) \Big],
\end{align*}
which implicitly treats all target tokens as equally important for task adaptation.
However, natural language generation admits substantial flexibility.
As observed in prior work~\citep{ruan2025enhancinglargelanguagemodel}, many tokens in reference responses are substitutable or stylistic, contributing little to core task reasoning.
Forcing the model to equally fit such tokens, especially when they are weakly supported by the instruction model, can introduce unnecessary noise and degrade stability after merging.

To address this issue, GIFT explicitly models token importance using guidance from the instruction-tuned model.
For each training example $(x,y)$, we perform a single forward pass of the instruction model to compute
\begin{equation}
q_t = p_{\theta_I}(y_t \mid x, y_{<t}), \quad t = 1, \ldots, T,
\end{equation}
where $T$ denotes the target sequence length.

We interpret $q_t$ as an importance score.
While prior approaches estimate data quality using a model's own likelihood~\citep{wu2025generalizationsftreinforcementlearning,zhu2026anchored}, the instruction-tuned model $\theta_I$ exhibits stronger alignment and task competence after post-training.
As a result, confidence signals derived from $\theta_I$ provide a more reliable calibration of which tokens reflect essential task knowledge (See Section \ref{sec:ablation}).

Since the dataset and instruction model are fixed, these importance scores are computed offline once per dataset, yielding augmented training tuples $(x,y,\mathbf{q})$, where $\mathbf{q} = [q_1, \ldots, q_T]$ is the sequence of confidence scores corresponding to each token in response $y$.
During fine-tuning, we incorporate token importance into the optimization objective:
\begin{align}
\mathcal{L}_{\mathrm{GIFT}}(\phi)
&= \mathbb{E}_{(x,y)\sim \mathcal{D}} \Big[ \sum_{t=1}^{T} q_t \, \ell_t(\phi) \Big],\\
\ell_t(\phi)
&= -\log p_{\theta_B,\phi}(y_t \mid x, y_{<t}),
\end{align}
where $q_t$ acts as an importance weight. In all experiments, we use $q_t$ directly as computed in Eq.~(3), without additional normalization, clipping, or temperature scaling. Tokens assigned low confidence by the instruction model contribute minimally to the gradient, while high-confidence tokens dominate the optimization signal. This mechanism concentrates learning on instruction-consistent and task-critical tokens, guiding the adapter toward instruction-aligned updates.

Unlike knowledge distillation, GIFT does not minimize a divergence between teacher and student distributions.
The instruction model is queried only once to produce scalar importance scores, which are then used to reweight the standard cross-entropy loss during base-model adapter training.
This design preserves the simplicity and efficiency of supervised fine-tuning while injecting guidance from the instruction model into the adaptation process.

\subsection{Transfer via Adapter Merge}

After optimization, the learned adapter $\phi^\star$ is merged into the instruction model:
\begin{equation}
\theta_I' = \theta_I + M(\phi^\star).
\end{equation}
Unlike post-hoc merging methods, the adapter learned by GIFT has been shaped throughout training by guidance.
As a result, the merged model inherits task-specific improvements while preserving the original instruction-following behavior and robustness.

We adopt standard LoRA merging in our experiments to minimize confounding factors and isolate the effect of guided fine-tuning. More generally, \textbf{GIFT is formulated as a unified pipeline for adapting instruction-tuned models, rather than a standalone training objective or a specialized merging procedure}. Its main contribution lies in integrating instruction-aware learning on the base model with a transfer mechanism that injects the resulting task-specific updates into the instruction-tuned model.

\section{Experiments}
\subsection{Experimental Setup}
\label{sec:experiment_setup}

We evaluate \textbf{GIFT} on both mathematical reasoning and knowledge-intensive question answering tasks to study whether guided adapter transfer can improve task performance under limited supervision. To ensure fair comparison, all methods use identical LoRA hyperparameters and are evaluated under the same decoding and evaluation protocols. Additional results on newer model families are provided in Appendix~\ref{sec:appendix_additional}.

\paragraph{Models.}
We consider both pretrained base models and their corresponding instruction-tuned variants. For mathematical reasoning, experiments are conducted on four model families: Llama3.1-8B and Llama3.2-3B~~\citep{llama3modelcard}, Qwen2.5-7B~~\citep{qwen2.5}, and DeepSeek-Math-7B~~\citep{shao2024deepseekmathpushinglimitsmathematical}.  
For knowledge-intensive evaluation, we follow prior work~~\citep{zhu2026anchored} and focus on the Llama3.1-8B family to study domain adaptation. 

\paragraph{Datasets.}
We evaluate \textbf{GIFT} on two representative domains: mathematical reasoning and knowledge-intensive question answering.  
For mathematical reasoning, we fine-tune models on 2,000 samples from NuminaMath-CoT~\citep{numina_math_datasets}, and evaluate performance on a diverse set of benchmarks, including Math500~\citep{hendrycks2021measuring}, Minerva Math~\citep{lewkowycz2022solving}, OlympiadBench~\citep{he2024olympiadbench}, AIME 2024~\citep{aime2024dataset}, and AMC 2023~\citep{amc2023dataset}. These benchmarks span varying difficulty levels and reasoning styles, providing a comprehensive assessment of mathematical generalization under limited supervision.
For knowledge-intensive evaluation, we fine-tune models on 10,000 samples from MedMCQA~\citep{pal2022medmcqa}. Evaluation is conducted on MMLU-medical~\citep{hendrycks2020measuring}, MedQA~\citep{jin2021disease}, and the MedMCQA test set, which collectively measure factual accuracy, domain knowledge coverage, and multiple-choice reasoning in the medical domain.

\begin{table*}[t]
\centering
\resizebox{0.8\textwidth}{!}{
\begin{tabular}{l|ccccc|c}
\toprule
\textbf{Methods} & \textbf{Math-500} & \textbf{Minerva Math} & \textbf{Olympiad Bench} & \textbf{AIME24} & \textbf{AMC23} & \textbf{Average} \\
\midrule
Llama3.1-8B-Instruct               & 37.4 & 16.8 & 9.8  & 1.9 & 18.0 & 16.8 \\ \midrule
\textit{w.} SFT                  & 23.0 & 7.7  & 4.9  & 0.6 & 8.3  & 8.9  \\
% \textit{w.} DFT                  & 44.2 & 21.3  & 12.8  & 3.7 & 19.7  & 20.4  \\
\textit{w.} Shadow-FT & 40.1 & 18.0 & 11.3 & 1.9 & 18.6 & 18.0 \\
\textit{w.} Re-Adapt          & 26.7 & 9.3  & 5.5  & 0.4 & 9.7  & 10.3 \\
\textit{w.} LoRE-Adapt             & 20.0 & 6.2  & 4.1  & 0.2 & 7.2  & 7.5  \\
\rowcolor{Highlight}
\textit{w.} \textbf{GIFT (Ours)}   & \textbf{45.8} & \textbf{19.9} & \textbf{15.9} & \textbf{5.0} & \textbf{23.3} & \textbf{22.0} \\
\bottomrule \toprule

Llama3.2-3B-Instruct               & 38.2 & 12.4 & 9.5  & 3.5 & 18.8 & 16.5 \\ \midrule
\textit{w.} SFT                  & 22.4 & 4.9  & 4.3  & 0.6 & 10.0 & 8.4  \\
% \textit{w.} DFT                  & 38.2 & 10.8  & 9.2  & 2.3 & 19.2  & 15.9  \\
\textit{w.} Shadow-FT & 37.6 & 12.1 & 10.7 & 4.2 & 18.9 & 16.7 \\
\textit{w.} Re-Adapt          & 19.0 & 4.1  & 3.9  & 0.4 & 9.7  & 7.4  \\
\textit{w.} LoRE-Adapt             & 9.6  & 2.4  & 2.1  & 0.2 & 3.9  & 3.6  \\
\rowcolor{Highlight}
\textit{w.} \textbf{GIFT (Ours)}   & \textbf{42.5} & \textbf{14.3} & \textbf{12.7} & \textbf{5.4} & \textbf{22.3} & \textbf{19.5} \\
\bottomrule \toprule

Qwen2.5-7B-Instruct                & 73.2 & 38.2 & 35.4 & 11.2 & 48.6 & 41.3 \\ \midrule
\textit{w.} SFT                   & 48.3 & 17.2 & 16.5 & 1.7  & 22.3 & 21.2 \\
% \textit{w.} DFT                  & 64.3 & 31.9  & 31.3  & 8.3 & 43.9  & 36.0  \\
\textit{w.} Shadow-FT & 70.6 & 36.6 & 31.3 & 7.9  & 43.4 & 38.0 \\
\textit{w.} Re-Adapt         & 68.9 & 33.4 & 30.1 & 10.4 & 44.5 & 37.5 \\
\textit{w.} LoRE-Adapt             & 66.1 & 27.3 & 29.0 & 7.7  & 41.1 & 34.2 \\
\rowcolor{Highlight}
\textit{w.} \textbf{GIFT (Ours)}   & \textbf{75.0} & \textbf{38.4} & \textbf{36.1} & \textbf{12.7} & \textbf{52.5} & \textbf{42.9} \\
\bottomrule \toprule

DeepSeek-Math-7B-Instruct          & 39.0 & 17.6 & 10.9 & 0.8 & 15.6 & 16.8 \\ \midrule
\textit{w.} SFT                   & 30.5 & 11.4 & 6.7  & 0.6 & 11.9 & 12.2 \\
% \textit{w.} DFT                  & 43.6 & 18.5  & 13.7  & 2.3 & 22.5  & 20.1  \\
\textit{w.} Shadow-FT & 39.9 & 18.3 & 11.7 & 1.0 & 16.2 & 17.4 \\
\textit{w.} Re-Adapt          & 33.6 & 13.3 & 8.2  & 0.6 & 15.0 & 14.1 \\
\textit{w.} LoRE-Adapt             & 31.6 & 12.4 & 7.6  & 0.2 & 11.7 & 12.7 \\
\rowcolor{Highlight}
\textit{w.} \textbf{GIFT (Ours)}   & \textbf{42.6} & \textbf{19.6} & \textbf{13.5} & \textbf{1.9} & \textbf{20.9} & \textbf{19.7} \\
\bottomrule
\end{tabular}
}
\caption{Average@16 accuracy of four large language models on mathematical reasoning benchmarks. The best performance of each model across benchmarks is bold. GIFT consistently surpasses all compared baselines across four model families, improving average accuracy while preserving general instruction-following robustness.}
\label{tab:math_results_gift}
\vspace{-6pt}
\end{table*}

\paragraph{Baselines.}
We compare GIFT with several representative adaptation and transfer baselines that differ in where task-specific updates are learned and how they are merged into the instruction-tuned model.

\noindent $\bullet$ \textbf{Instruct Model}. 
The original instruction-tuned checkpoint released by the model developers serves as both a strong baseline and the merge target for all transfer-based methods.

\noindent $\bullet$ \textbf{Instruct-SFT}. 
Direct supervised fine-tuning of the instruction-tuned model.

\noindent $\bullet$ \textbf{Shadow-FT / Chat Vector}~\citep{huang-etal-2024-chat,wu2025shadowfttuninginstructmodel}. 
Both methods fine-tune the base model and then directly merge the learned weight updates or adapters into the instruction-tuned model. 

\noindent $\bullet$ \textbf{Re-Adapt / Task Arithmetic}~\citep{fleshman2024readaptreverseengineeredadaptation,ilharco2023editing}. 
Linear combinations of the base model, instruction vector, and task-adapted parameters, using coefficients of $0.5$ as recommended~\citep{fleshman2024readaptreverseengineeredadaptation,ilharco2023editing}.

\noindent $\bullet$ \textbf{LoRE-Adapt}~\citep{fleshman2024readaptreverseengineeredadaptation}. 
A low-rank variant of Re-Adapt that applies truncated SVD to the instruction offset, yielding a compact instruction adapter.

\paragraph{Training Details.}
All methods are trained using the AdamW optimizer for 1 epoch with a maximum sequence length of 2048 tokens.
To ensure a controlled comparison, we use an identical LoRA configuration across all experiments: global batch size of 256, learning rate $2\times10^{-4}$, LoRA rank $r=64$, LoRA scaling $\alpha=128$, LoRA dropout $0.05$, and warmup ratio $0.1$.
Unless otherwise specified, we keep the remaining optimization settings fixed across all experiments.
For mathematical reasoning benchmarks, following previous works~~\citep{wu2025generalizationsftreinforcementlearning}, we use stochastic decoding with temperature $1.0$ and maximum generation length $4096$. We sample $16$ responses per problem and report Average@16, i.e., the average accuracy across these $16$ stochastic samples. This setting maintains comparable accuracy while also reflecting the model's exploration ability under test-time sampling. 
For medical, we use standard multiple-choice accuracy.

\subsection{Main Results}
\label{sec:results}

\begin{table}[t]
\centering
\resizebox{\linewidth}{!}{
\begin{tabular}{l|ccc|c}
\toprule
\textbf{Methods} & \textbf{MedQA} & \textbf{MMLU-medical} & \textbf{MedMCQA} & \textbf{Average} \\
\midrule
Llama3.1-8B-Instruct              & 55.2 & 75.1 & 57.4 & 62.6 \\
\textit{w.} SFT                   & 50.0 & 64.0 & 57.9 & 57.3 \\
\textit{w.} Shadow-FT & 65.6 & 73.8 & 55.9 & 65.1 \\
\textit{w.} Re-Adapt         & 63.6 & 71.4 & 54.2 & 63.1 \\
\textit{w.} LoRE-Adapt            & 62.6 & 70.1 & 54.3 & 62.4 \\
\rowcolor{Highlight}
\textbf{w. GIFT (Ours)}           & \textbf{68.3} & \textbf{77.7} & \textbf{60.2} & \textbf{68.8} \\
\bottomrule
\end{tabular}
}
\caption{Results on medical QA benchmarks. GIFT improves knowledge acquisition while maintaining the general capabilities of the instruction model.}
\label{tab:gift_medical}
\vspace{-6pt}
\end{table}

\begin{table*}[!htb]
\centering
\resizebox{0.8\linewidth}{!}{
\begin{tabular}{lcccccc}
\toprule
\textbf{Ablation} 
& \textbf{Math-500} & \textbf{Minerva Math} & \textbf{Olympiad Bench} & \textbf{AIME24} & \textbf{AMC23} & \textbf{Average} \\
\midrule
Instruct Model      & 73.2 & 38.2 & 35.4 & 11.2 & 48.6 & 41.3 \\
SFT+Merge           & 70.6 & 36.6 & 31.3 & 7.9 & 43.4 & 38.0 \\
GIFT-BaseT          & 73.5 & 37.3 & 34.6 & 12.1 & 47.7 & 41.0 \\
\textbf{GIFT (ours)}& 75.0 & 38.4 & 36.1 & 12.7 & 52.5 & 42.9 \\
\bottomrule
\end{tabular}}
\caption{Ablation on Qwen2.5-7B. We compare the original instruction-tuned model, standard SFT with merge, a GIFT variant
that uses the base model as teacher (GIFT-BaseT), and the full GIFT method that uses the instruction model as teacher.}
\label{tab:ablation_qwen}
\vspace{-6pt}
\end{table*}

From Table~\ref{tab:math_results_gift} and Table~\ref{tab:gift_medical}, we draw the following observations.

\textbf{GIFT achieves consistent and substantial performance gains across models and benchmarks.}
Across five instruction-tuned backbones and all evaluated tasks, GIFT consistently improves average accuracy over both the original instruction models and prior adaptation methods. On mathematical reasoning, GIFT improves the average score by +5.2 on Llama3.1-8B, +3.0 on Llama3.2-3B, +2.9 on DeepSeek-Math-7B, and +1.6 on Qwen2.5-7B, demonstrating stable gains across model scales. In contrast, direct fine-tuning of instruction models (Instruct-SFT) leads to substantial degradation, particularly under limited supervision and stochastic decoding with temperature $1.0$, which amplifies distributional shifts in instruction-tuned checkpoints. By learning task-specific updates on the base model and transferring them under instruction guidance, GIFT yields reliable improvements across all benchmarks.

\textbf{GIFT remains effective even on strong instruction-tuned models.}
Improving highly optimized models such as Qwen2.5-7B-Instruct is particularly challenging. While Shadow-FT provide moderate gains on relatively weaker backbones (e.g., Llama and DeepSeek-Math), their performance degrades on Qwen2.5-7B-Instruct. In contrast, GIFT consistently delivers further improvements on this strong baseline, indicating that instruction-guided fine-tuning can introduce task-specific knowledge without disrupting existing capabilities.

\textbf{GIFT generalizes across model families and domains.}
Beyond mathematical reasoning, GIFT also transfers effectively to knowledge-intensive medical QA. As shown in Table~\ref{tab:gift_medical}, GIFT improves the average score on Llama3.1-8B-Instruct from 62.6 to 68.8 (+6.2), achieving the best performance across MedQA, MMLU-medical, and MedMCQA. This result demonstrates that confidence-guided adaptation is not limited to reasoning-centric tasks, but also benefits fact-heavy domains that require accurate knowledge acquisition and structured decision-making.

\subsection{Ablation}
\label{sec:ablation}

Table~\ref{tab:ablation_qwen} presents an ablation study on Qwen2.5-7B, a strong and highly optimized instruction-tuned model on which further improvements are known to be challenging. The goal of this study is to isolate the contribution of guidance from the instruction model in the proposed GIFT framework.

We consider three representative adaptation settings. 
(1) \textbf{SFT+Merge} corresponds to the Shadow-FT setting described in Section~\ref{sec:experiment_setup}, where a low-rank adapter is trained on the pretrained base model using standard supervised fine-tuning and then merged into the instruction-tuned model.
(2) \textbf{GIFT-BaseT} adopts the same guided optimization framework as GIFT, but uses the base model itself as the teacher, allowing us to control for the effect of reweighting without instruction-aligned guidance. 
(3) \textbf{GIFT} uses the instruction-tuned model as the teacher during adapter training, and merges the resulting adapter back into the instruction model.

As shown in Table~\ref{tab:ablation_qwen}, directly merging an adapter trained via standard SFT on the base model (\textbf{SFT+Merge}) leads to a clear performance degradation compared to the original instruction-tuned model, with an average drop of 3.3 points. This result indicates that unguided task adaptation can introduce parameter updates that are incompatible with the instruction-tuned representation space, even when merged post hoc.
Using the base model as the teacher (\textbf{GIFT-BaseT}) largely recovers the instruction baseline, achieving an average score comparable to the original instruction model (41.0 vs.\ 41.3), but fails to yield consistent improvements across benchmarks. This suggests that reweighting alone is insufficient when the guidance signal does not reflect instruction-aligned behavior.
In contrast, the full \textbf{GIFT} method, which leverages guidance from the instruction-tuned model, consistently improves performance across all benchmarks, achieving a +1.6 average gain over the instruction baseline. This comparison highlights that the effectiveness of GIFT crucially depends on instruction-aligned guidance during fine-tuning, rather than merely reweighting tokens or transferring task-specific updates.

Overall, these results confirm that incorporating the instruction-tuned model during the fine-tuning stage is essential for learning adapters that are both merge-compatible and performance-improving.

\begin{figure*}[t]
  \includegraphics[width=\linewidth]{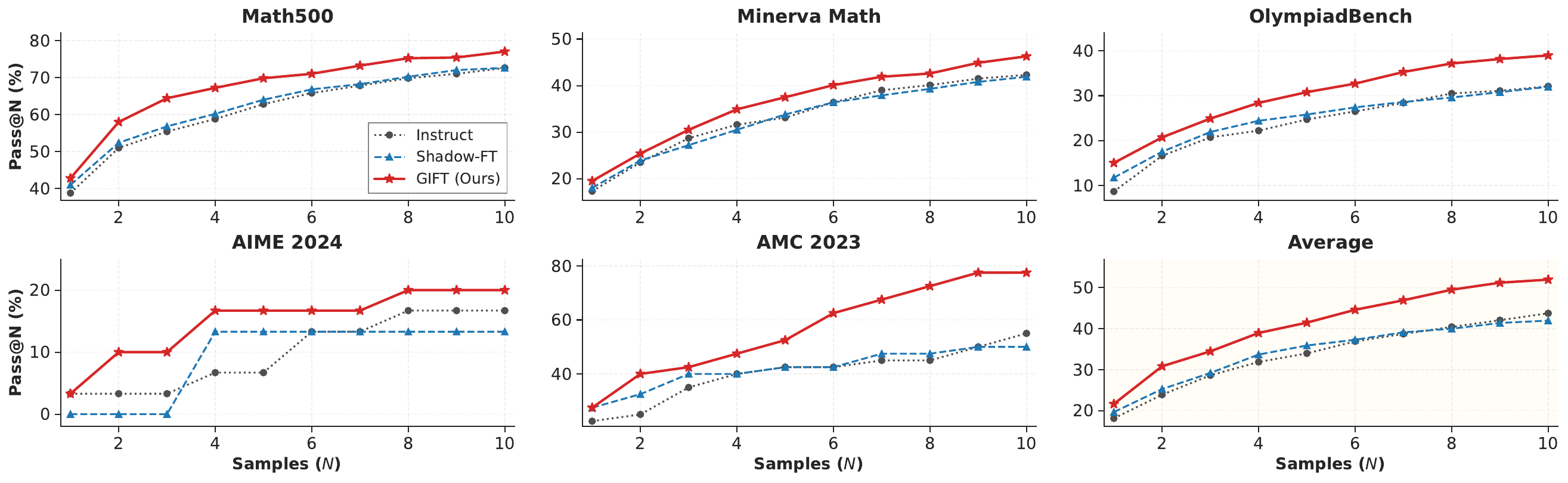}
  \caption{Pass@$N$ performance under test-time scaling on Llama3.1-8B across five mathematical benchmarks and their average. GIFT consistently outperforms the instruction baseline and Shadow-FT at all sampling budgets.}
  \label{fig:test-time-scaling}
\end{figure*}

\section{Analyses}

\subsection{Test-Time Scaling}
\label{sec:test_time_scaling}

Inference-time strategies such as Best-of-$N$ (BoN) sampling \citep{charniak-johnson-2005-coarse,NEURIPS2020_1f89885d}, which generate multiple candidate outputs for each query, are widely adopted to enhance LLM reasoning \citep{welleck2024from,snell2024scalingllmtesttimecompute,ruan-etal-2025-g2}. The effectiveness of these methods relies on whether the generated candidates can sufficiently explore the solution space.  

To quantify this effect, we adopt the \textbf{Pass@$N$} metric, which can be regarded as a special case of BoN \citep{chen2021evaluatinglargelanguagemodels}. For a question $q$, let $\{o_i\}_{i=1}^N$ denote $N$ outputs and $\mathcal{F}(o_i)$ a verifier returning $1$ if $o_i$ is correct and $0$ otherwise. Then,  
\begin{equation}
\small
\mathrm{Pass}@N(q) = \mathbb{I}\!\Big[\exists\, i \in \{1,\ldots,N\} \ \text{s.t.}\ \mathcal{F}(o_i)=1\Big].
\end{equation}
This metric evaluates whether at least one correct solution is found among $N$ attempts.  

Figure~\ref{fig:test-time-scaling} reports the Pass@$N$ performance of the instruction baseline, Shadow-FT, and GIFT on Llama3.1-8B across five mathematical benchmarks and their average.
Across all tasks, \textbf{GIFT consistently outperforms both baselines at every sampling budget}. The performance gap is already present at $N=1$ and remains stable as $N$ increases, indicating that the gains introduced by GIFT are not restricted to single-sample accuracy.
On more challenging benchmarks such as OlympiadBench and AIME 2024, GIFT exhibits stronger scaling behavior, accumulating correct solutions more effectively as additional samples are generated. In contrast, Shadow-FT shows limited or saturated improvements under larger sampling budgets.
Overall, these results suggest that guided fine-tuning improves not only base accuracy but also the effectiveness with which models exploit test-time sampling, producing adaptations that are more compatible with inference-time scaling.

\subsection{Impact on General Capabilities}
\label{sec:generalization_ood}

\begin{table}[!htb]
\centering
\resizebox{0.9\linewidth}{!}{
\begin{tabular}{l|cc}
\toprule
\textbf{Model} & \textbf{MMLU} & \textbf{IFEval} \\
\midrule
Qwen2.5-7B-Instruct        & 68.7 & 71.2 \\
\textit{w.} Shadow-FT     & 68.6 & 71.9 \\
\textit{w.} GIFT & 68.8 & 72.1 \\
\midrule
Llama3.1-8B-Instruct      & 63.2 & 73.8 \\
\textit{w.} Shadow-FT     & 63.7 & 73.6 \\
\textit{w.} GIFT & 63.7 & 74.7 \\
\bottomrule
\end{tabular}
}
\caption{Impact of merging task-adapted adapters on general knowledge (MMLU) and instruction-following (IFEval) tasks.}
\label{tab:ood_eval}
\vspace{-6pt}
\end{table}

Beyond task-specific improvements, we examine whether merging task-adapted adapters affects the general capabilities of instruction-tuned models.
We evaluate merged models on two complementary benchmarks:
MMLU~\citep{hendrycks2020measuring}, which measures broad multi-domain knowledge, and IFEval~\citep{zhou2023instructionfollowing}, which evaluates instruction-following robustness under strict prompt-level constraints.

Table~\ref{tab:ood_eval} reports results on Qwen2.5-7B-Instruct and Llama3.1-8B-Instruct.
Across both backbones, GIFT preserves the original instruction model’s performance on MMLU, with results matching or slightly exceeding the instruction baseline.
This indicates that guided fine-tuning and merging do not compromise general knowledge coverage.

On IFEval, GIFT maintains or slightly improves prompt-level strict accuracy after merging, compared to both the instruction baseline and Shadow-FT.
Across both backbones, no degradation in instruction-following robustness is observed.

Overall, these results indicate that merging GIFT-trained adapters does not compromise the general capabilities of instruction-tuned models.
The preserved performance on MMLU and IFEval suggests that GIFT enables task adaptation while remaining largely neutral to the model’s general knowledge and instruction-following behavior.

\subsection{Generalization to Instruction Tasks}
\label{sec:sni_generalization}

To further examine whether GIFT generalizes beyond math and medical QA, we construct an additional instruction-following evaluation from summarization-style tasks in Super-NaturalInstructions~\citep{supernaturalinstructions}.
We use the official default split, filter tasks by keywords related to summarization, sample 2,000 training examples, and evaluate on 100 test instances per task over 8 tasks (800 examples in total).
Table~\ref{tab:app_sni_results} shows that GIFT improves both exact match and RougeL over the original instruction model and Shadow-FT.
These results suggest that guided base-to-instruct transfer remains effective on a qualitatively different task family involving multi-token generation.

\begin{table}[!htb]
\centering
\resizebox{0.8\linewidth}{!}{
\begin{tabular}{l|cc}
\toprule
\textbf{Model} & \textbf{EM} & \textbf{RougeL} \\
\midrule
Llama3.1-8B-Instruct & 10.75 & 37.38 \\
\textit{w.} Shadow-FT & 9.50 & 37.65 \\
\textit{w.} GIFT & \textbf{12.00} & \textbf{40.28} \\
\bottomrule
\end{tabular}
}
\caption{Results on summarization-style tasks from Super-NaturalInstructions.}
\label{tab:app_sni_results}
\vspace{-6pt}
\end{table}

\subsection{Model Size Scaling}
\label{sec:model_size_scaling}

\begin{table}[!htb]
\centering
\resizebox{\linewidth}{!}{
\begin{tabular}{l|c|c|c}
\toprule
\textbf{Model Size} & \textbf{Instruct} & \textbf{Shadow-FT} & \textbf{GIFT} \\
\midrule
Qwen2.5-0.5B  & 7.9  & 2.8  & \textbf{8.3} \\
Qwen2.5-1.5B  & 18.3 & 9.8  & \textbf{22.0} \\
Qwen2.5-3B    & 30.4 & 24.5 & \textbf{32.8} \\
Qwen2.5-7B    & 41.3 & 38.0 & \textbf{42.9} \\
Qwen2.5-14B   & 46.9 & 44.9 & \textbf{48.1} \\
Qwen2.5-32B   & 50.6 & 48.2 & \textbf{51.2} \\
\bottomrule
\end{tabular}
}
\caption{Average accuracy across five mathematical reasoning benchmarks for the Qwen2.5 family across different scales.}
\label{tab:qwen_scaling_avg}
\end{table}

We further examine whether the gains of GIFT persist across different model scales. 
To this end, we conduct a model size scaling study on the Qwen2.5 family, which spans from 0.5B to 32B parameters under a unified architecture and training recipe. 
This setting allows us to analyze how guided adapter transfer interacts with model capacity in a controlled manner.

Table~\ref{tab:qwen_scaling_avg} reports the average accuracy across five mathematical reasoning benchmarks for each model size; full per-benchmark results are provided in Table~\ref{tab:app_qwen_scaling_full}.
Across all scales, GIFT consistently outperforms both the instruction-tuned baseline and Shadow-FT, demonstrating stable improvements from small (0.5B) to large (32B) models.

Notably, the performance gap between GIFT and the baselines does not diminish as model size increases. 
While all methods benefit from larger model capacity, guided adapter transfer provides complementary gains that persist even for strong instruction-tuned models such as Qwen2.5-14B and Qwen2.5-32B. 
In contrast, Shadow-FT exhibits weaker or inconsistent improvements, particularly for smaller and mid-sized models, where unguided merging is more susceptible to instability.

Overall, these results indicate that GIFT scales favorably with model size and remains effective across a wide range of capacities, reinforcing its applicability as a general-purpose adaptation strategy for instruction-tuned language models.

\subsection{Learning Signal Redistribution}
\label{sec:token_analysis}

\begin{figure}[t]
    \centering
    \includegraphics[width=0.9\linewidth]{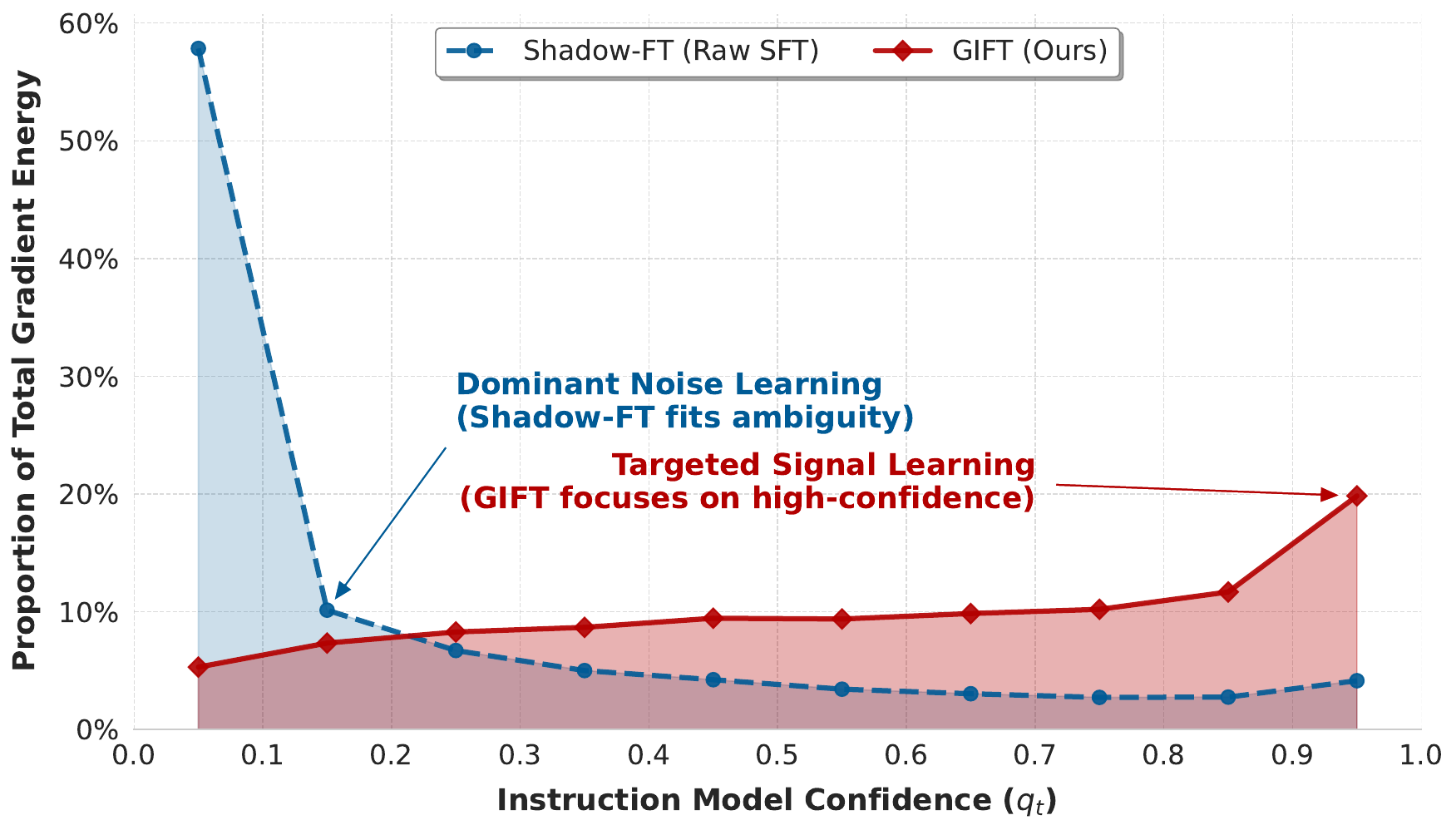}
    \caption{Training signal distribution across instruction-model confidence levels, where cross-entropy losses in GIFT are weighted by confidence $q_t$.}
    \label{fig:gift_mechanism_analysis}
\end{figure}

We analyze how guided fine-tuning redistributes training signal across tokens with different instruction-model confidence levels.
Following common practice, we use cross-entropy loss as a proxy for learning signal magnitude, where higher loss typically induces stronger parameter updates.

Our analysis covers over 500k target tokens from the NuminaMath-CoT training set using Llama3.1-8B.
For each token, we record its cross-entropy loss $\ell_t$ and instruction-model confidence $q_t$, and group tokens into bins from low to high confidence.
We compare Shadow-FT~\cite{wu2025shadowfttuninginstructmodel}, where the learning signal is proportional to $\ell_t$, with \textbf{GIFT}, where it is proportional to $q_t \cdot \ell_t$.

Figure~\ref{fig:gift_mechanism_analysis} shows the normalized contribution of each confidence region.
Under Shadow-FT, 79.7\% of the learning signal originates from low-confidence tokens, which also exhibit the highest average loss, indicating that standard fine-tuning emphasizes uncertain or ambiguous regions.
In contrast, GIFT substantially reshapes this distribution.
The contribution from low-confidence tokens drops to 29.6\%, while medium- and high-confidence regions receive a much larger share.
Notably, high-confidence tokens contribute only 6.9\% of the signal under Shadow-FT but 31.5\% under GIFT, despite having lower raw losses.

Overall, GIFT selectively redistributes learning signals rather than uniformly amplifying gradients.
By suppressing updates from instruction-uncertain tokens and prioritizing instruction-consistent regions, GIFT steers optimization toward more compatible update directions, providing an empirical explanation for its improved stability and merge compatibility.

\subsection{Offline Annotation Cost}
\label{sec:annotation_cost}

Since GIFT requires a one-time offline pass over the training set to compute $q_t$, we additionally profile this preprocessing step.
Table~\ref{tab:annotation_cost} summarizes the profiling results for the main mathematical setup with 2,000 NuminaMath-CoT samples on a single RTX 4090 24GB GPU using batch size $1$.
These results support the claim that the annotation stage is lightweight in our setting, although larger datasets would still incur additional preprocessing cost.

\begin{table}[t]
\centering
\resizebox{0.95\linewidth}{!}{
\begin{tabular}{lccc}
\toprule
\textbf{Teacher Model} & \textbf{Time} & \textbf{Peak Mem.} & \textbf{File Size} \\
\midrule
Llama3.1-8B-Instruct & 2m11s & $<22$GB & 11.91MB \\
Qwen2.5-7B-Instruct & 2m9s & $<22$GB & 11.75MB \\
\bottomrule
\end{tabular}
}
\caption{Offline annotation cost for computing token-level confidence scores on 2,000 NuminaMath-CoT samples. The raw 2K file size is 2.9MB.}
\label{tab:annotation_cost}
\vspace{-6pt}
\end{table}

\section{Conclusion}
We introduced \textbf{GIFT}, a guided fine-tuning and transfer framework for adapting instruction-tuned language models under limited supervision. By leveraging effective confidence signals from the instruction model to guide adapter training on the base model, GIFT enables more stable and effective acquisition of task-specific knowledge. Extensive experiments on mathematical reasoning and knowledge-intensive benchmarks show that GIFT consistently outperforms direct fine-tuning and existing merge-based methods, while preserving the general instruction-following behavior of instruction-tuned models.

% \clearpage
% \newpage 

\section*{Limitations}
GIFT requires an offline annotation step to compute confidence scores for the training dataset.
Although this process is lightweight and performed only once, it introduces additional preprocessing cost for very large datasets.
Exploring more efficient or on-the-fly approximations of guidance from the instruction model is a promising avenue for future research.
While we focus on standard adapter merging for clarity, future work could explore combining GIFT with advanced merging techniques such as Fisher-weighted averaging or TIES.
We leave this direction as future work, as it is orthogonal to the core contribution of guided fine-tuning.

\bibliography{custom}

\appendix

\section{Supplementary Experimental Results}
\label{sec:appendix}

Table~\ref{tab:app_qwen_scaling_full} reports the full per-benchmark results corresponding to the model size scaling study in Section~\ref{sec:model_size_scaling}. 
We include detailed accuracies on each mathematical benchmark to complement the averaged results in Table~\ref{tab:qwen_scaling_avg} and facilitate more fine-grained comparison across model scales and methods.

\begin{table*}[!htb]
\centering
\resizebox{0.8\textwidth}{!}{
\begin{tabular}{l|ccccc|c}
\toprule
Model & Math500 & Minerva & Olympiad & AIME24 & AMC23 & Avg \\
\midrule
Qwen2.5-0.5B-Instruct & 22.9 & 3.1 & 4.2 & 0.6 & 8.6 & 7.9 \\
w. Shadow-FT & 9.5 & 1.2 & 1.5 & 0.0 & 1.7 & 2.8 \\
w. GIFT & 25.2 & 3.7 & 5.4 & 0.4 & 6.9 & 8.3 \\
\midrule
Qwen2.5-1.5B-Instruct & 44.4 & 11.0 & 13.6 & 0.6 & 21.7 & 18.3 \\
w. Shadow-FT & 26.6 & 5.8 & 6.8 & 0.8 & 8.8 & 9.8 \\
w. GIFT & 51.1 & 12.8 & 16.5 & 2.3 & 27.5 & 22.0 \\
\midrule
Qwen2.5-3B-Instruct & 62.9 & 23.3 & 24.3 & 5.2 & 36.1 & 30.4 \\
w. Shadow-FT & 54.7 & 18.6 & 18.6 & 3.1 & 27.3 & 24.5 \\
w. GIFT & 65.8 & 27.5 & 26.2 & 4.0 & 40.6 & 32.8 \\
\midrule
Qwen2.5-7B-Instruct & 73.2 & 38.2 & 35.4 & 11.2 & 48.6 & 41.3 \\
w. Shadow-FT & 70.6 & 36.6 & 31.3 & 7.9 & 43.4 & 38.0 \\
w. GIFT & 75.0 & 38.4 & 36.1 & 12.7 & 52.5 & 42.9 \\
\midrule
Qwen2.5-14B-Instruct & 77.5 & 44.5 & 40.0 & 14.2 & 58.1 & 46.9 \\
w. Shadow-FT & 76.9 & 43.8 & 38.1 & 10.8 & 54.8 & 44.9 \\
w. GIFT & 78.9 & 47.4 & 41.1 & 14.0 & 59.1 & 48.1 \\
\midrule
Qwen2.5-32B-Instruct & 81.3 & 47.0 & 44.3 & 16.7 & 63.6 & 50.6 \\
w. Shadow-FT & 79.8 & 40.1 & 43.0 & 14.2 & 64.1 & 48.2 \\
w. GIFT & 81.5 & 48.6 & 44.8 & 16.0 & 65.2 & 51.2 \\
\bottomrule
\end{tabular}
}
\caption{Full results on five mathematical reasoning benchmarks for Qwen2.5 models at different scales.}
\label{tab:app_qwen_scaling_full}
\vspace{-6pt}
\end{table*}

\section{Additional Experiments}
\label{sec:appendix_additional}

\subsection{Results on Qwen3-8B}

We additionally evaluate GIFT on Qwen3-8B under the same NuminaMath-CoT training setup used in the main mathematical experiments.
At evaluation time, rather than relying on training-free techniques \citep{laibeyond, huang2026satbalancingreasoningaccuracy}, we use the no-thinking mode by setting \texttt{enable\_thinking=False} in the chat template, since the training responses are standard chain-of-thought solutions rather than explicit thinking-mode traces.
Table~\ref{tab:app_qwen3_results} shows that GIFT remains effective on this newer model family, improving the Qwen3-8B baseline by 1.2 average points and outperforming Shadow-FT by 3.2 points.

\begin{table*}[!htb]
\centering
\resizebox{0.8\textwidth}{!}{
\begin{tabular}{l|ccccc|c}
\toprule
\textbf{Method} & \textbf{Math-500} & \textbf{Minerva Math} & \textbf{Olympiad Bench} & \textbf{AIME24} & \textbf{AMC23} & \textbf{Average} \\
\midrule
Qwen3-8B & 82.5 & \textbf{37.6} & 47.1 & 24.8 & 67.2 & 51.8 \\
\textit{w.} Shadow-FT & 81.2 & 32.2 & 45.1 & 24.4 & 65.9 & 49.8 \\
\textit{w.} GIFT & \textbf{83.2} & 36.9 & \textbf{48.2} & \textbf{26.7} & \textbf{70.2} & \textbf{53.0} \\
\bottomrule
\end{tabular}
}
\caption{Additional mathematical reasoning results on Qwen3-8B.}
\label{tab:app_qwen3_results}
\vspace{-6pt}
\end{table*}

\subsection{Confidence Token Analysis}

To better understand why instruction-model confidence can serve as a useful weighting signal, we compare token-type shares in the full NuminaMath-CoT training responses and in the high-confidence subset with $q_t \ge 0.9999$.
We use digits as a simple proxy for math-critical content.
Digits are consistently enriched among high-confidence tokens for both teachers: for Llama3.1-8B-Instruct, digits account for 11.35\% of all response tokens but 20.69\% of the high-confidence subset (1.82$\times$); for Qwen2.5-7B-Instruct, the corresponding values are 14.49\% and 23.27\% (1.61$\times$).
This pattern suggests that instruction-derived confidence tends to upweight more structured, numerically grounded portions of mathematical solutions.

\subsection{Confidence Across Difficulty Levels}

To connect Figure~\ref{fig:gift_mechanism_analysis} with benchmark difficulty, we additionally compute token-level confidence statistics on reference solutions from an easier set (GSM8K) and a harder set (AIME24).
Across both teacher models, AIME24 consistently contains a larger share of low-confidence regions.
With Llama3.1-8B-Instruct as teacher, GSM8K has $q_{\text{mean}}=0.748$, $\mathrm{frac}(q_t \ge 0.9)=60.58\%$, and $\mathrm{frac}(q_t < 0.5)=23.86\%$, whereas AIME24 has $q_{\text{mean}}=0.652$, $\mathrm{frac}(q_t \ge 0.9)=45.19\%$, and $\mathrm{frac}(q_t < 0.5)=34.04\%$.
With Qwen2.5-7B-Instruct as teacher, GSM8K has $q_{\text{mean}}=0.816$, $\mathrm{frac}(q_t \ge 0.9)=74.08\%$, and $\mathrm{frac}(q_t < 0.5)=17.80\%$, while AIME24 has $q_{\text{mean}}=0.690$, $\mathrm{frac}(q_t \ge 0.9)=52.04\%$, and $\mathrm{frac}(q_t < 0.5)=30.07\%$.
These statistics support the intuition that harder reasoning benchmarks expose more instruction-uncertain regions, where GIFT suppresses unstable learning signals.

\subsection{Prompt Formatting Details}

For completeness, we summarize the prompt formatting used in our experiments.
For NuminaMath-CoT, we apply the chat template with \texttt{add\_generation\_prompt=True} and append the instruction line ``Please reason step by step, and put your final answer within \verb|\boxed{}|.'' to the user message.
The training target is the provided solution response, which includes both the reasoning process and the boxed final answer.

For MedMCQA, we format each example as a multiple-choice prompt with options in the standard ``A. / B. / C. / D.'' format, then append the instruction ``Please reason step by step. At the end of your response, you MUST conclude with the exact phrase: `So the answer to this question is [Option]'.''
This formatting makes answer extraction unambiguous during evaluation.

For the additional summarization experiment in Super-NaturalInstructions, each instance is serialized as \texttt{Task definition: <Definition>\textbackslash n\textbackslash n Input: <input>\textbackslash n\textbackslash n Output:}, and the first reference output is used as the response target.

\section{LLM Usage}
In the preparation of this paper, we only used large language models (LLMs) as an assistive tool for grammar correction and text polishing.

\end{document}